\documentclass[11pt]{article} 

\usepackage{graphicx}
\usepackage{booktabs}
\usepackage{algorithm}
\usepackage{algpseudocode}
\usepackage[T1]{fontenc}
\usepackage{lmodern}
\usepackage[flushleft]{threeparttable}
\usepackage[hidelinks]{hyperref} %
\hypersetup{hidelinks}
\usepackage{multirow}
\usepackage{longtable}
\usepackage[margin=3cm]{geometry} %
\usepackage{amsmath}
\usepackage{authblk}
\usepackage{caption}
\usepackage{subcaption}
\usepackage{listings}
\usepackage{float}
\usepackage{array}
\usepackage{makecell}

\title{Cavity Duplexer Tuning with 1d Resnet-like Neural Networks}
\author[1]{Anton Raskovalov}
\affil[1]{Joint Stock "Research and production company "Kryptonite" \authorcr
E-mail: a.raskovalov@kryptonite.ru}
\date{}
\begin{document}

    \captionsetup[table]{labelformat={default},labelsep=period,name={Table}}

    \maketitle

    \begin{abstract}
        This paper presents machine learning method for tuning of cavity duplexer with a large amount of adjustment screws. After testing we declined conventional reinforcement learning approach and reformulated our task in the supervised learning setup. The suggested neural network architecture includes 1d ResNet-like backbone and processing of some additional information about S-parameters, like the shape of curve and
peaks positions and amplitudes. This neural network with external control algorithm is capable to reach almost the tuned state of the duplexer within 4-5 rotations per screw.
    \end{abstract}

    
    \section{Introduction}
\label{intro}

In wireless communications it is important to filter or combine/separate signals in a specific range of frequencies. Such devices are called filters and duplexers, respectively. In the microwave domain they use phenomena of electromagnetic resonance in circuits that include multiple capacitances and inductances created with a number of coupled cavity resonators to pass or reject specific frequencies and therefore referred as ``cavity'' filters/duplexers. They are manufactured as metallic boxes with few connectors (2 for a filter and 3 for a duplexer). Due to small imperfections in the assembly of the enclosure, the frequency response varies from one device to another. The frequency responses of such devices are called the scattering parameters or S-parameters. The cavity filter/duplexer housing includes many tuning screws that allow adjustment of the S-parameters to meet specific requirements. The tuning process is very labor-intensive and takes several hours per single device even for experienced personnel.

It has recently been proposed that machine learning techniques can be used to assist with duplexer tuning. Unfortunately, existing studies are scarce and typically address only simple, model setups. Michalski~\cite{1} used simple feed forward neural network with one hidden layer of 50 neurons to tune 6- or 11-cavity filter. Wang and Ou~\cite{2} applied more robust technique, reinforcement learning (RL), to tune only 4 screws. Their neural networks consisted of 2 hidden layers with 300 and 600 neurons and used a 20-dimensional feature set obtained with principal component analysis as input instead of S-parameters. Aghanim et al.~\cite{3} also used reinforcement learning and reduced input dimensionality with principal component analysis to tune just 2 screw-filter. From~7 to~13 screws filters were studied in Master's thesis of Larsson ~\cite{4}. As one can see all these studies address toy problems, while devices of interest have more than several dozen screws.

In this work we will show how it is possible to solve this task in terms of supervised learning without using RL methodology. This paper is organized as follows: Section~\ref{metho} describes the suggested methodology and architectures, Section~\ref{results} presents the research results and Section~\ref{concl} summarizes the conclusions.

\section{Methodology}
\label{metho}

\subsection{Terms and Metrics}
\label{terms}

Our aim is to tune a cavity duplexer which has two passbands and three S-parameters (S-curves): S11, S21, and S31. These curves we obtain from spectrum analyzer as a sequence of amplitudes in a given frequency range. After selecting the frequency band of interest, we obtain 1300~points per one S-curve. These three 1300-dimensional vectors are referred to as a ``state'' of duplexer or ``state-tensor''. The tuning process is carried out by rotating the tuning screws in enclosure of the filter. Our duplexer contains several dozen screws, so we need to predict several dozen separate values, describing screw rotation, for tuning each screw. Following RL terminology, we call this vector of values the ``action'' or ``action-vector''. A tuned duplexer must satisfy the amplitude specifications over the designated frequency bands. Here, we assume that the S11 curve needs to be below -20~dB in both passbands. There are also requirements for S21 and S31, but they are usually met automatically when S11 satisfies the specified conditions and we will not consider them here.

\begin{figure}[b]
	\centering
	\includegraphics[width=\linewidth]{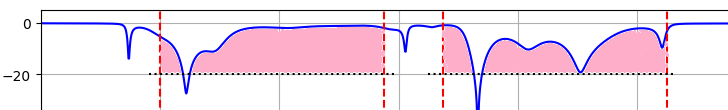} 
	\caption{An example of S11 curve, two passbands are bounded by vertical dash lines. The discussed metrics, areas (formula \protect\eqref{eq:1}), are colored pink in both passbands. For the tuned cavity filter these areas must be equal to zero.}
	\label{fig:1}
\end{figure}

As a measure of the quality of filter tuning, we will consider an area between S11-curve and the horizontal line -20~dB in each passband (Figure~\ref{fig:1}). These metrics (referred as an ``area'') can be calculated using the following equation:

\begin{equation}
\text{area} = \sum \text{ReLU}(S_{11}[\text{passband}] + 20), \label{eq:1}
\end{equation}

where $\text{ReLU}(x) = \max(0, x)$, $[\text{passband}]$ is the range of corresponding indices. Typical value for untuned duplexer of both areas (for each passband) is about 4000-6000~units. For a fully tuned filter these areas must be equal to zero. Our aim is to develop an artificial neural network (NN) which takes the state-tensor as an input and returns the action-vector which contains the values by which duplexer screws should be rotated to reduce both areas to zero. This neural network we will refer to as ``actor''.

As a physical duplexer can be damaged during huge number of experiments, we work with a special software which mimics its behavior. We refer to this software as a ``simulator'' or an ``environment''. In this paper all duplexer response data are obtained using this simulator. For cases when we work with data obtained from simulator and saved in specific datasets, we will use a term ``offline'', otherwise for direct interactions with the environment we will use the term ``online''.

\subsection{From RL to supervised learning}
\label{from}

As shown in Section~\ref{intro}, RL is usually used for cavity filter tuning. However, RL suffers from several disadvantages, which typically include instability during the training~\cite{5,6}, difficulties in hyperparameter tuning~\cite{7,8}, ambiguity in a reward function choice~\cite{9,10}, the challenge of finding an appropriate balance between exploration and exploitation~\cite{11}. However, RL focuses on maximizing reward along some trajectory (path). It is relevant for certain tasks as automotive or cleaner robots, but in fact does not matter for our purposes. We are not interested in tuning path and only want to reach a final state, a tuned duplexer, as soon as possible.

\begin{figure}[h]
	\centering
	\includegraphics[width=\linewidth]{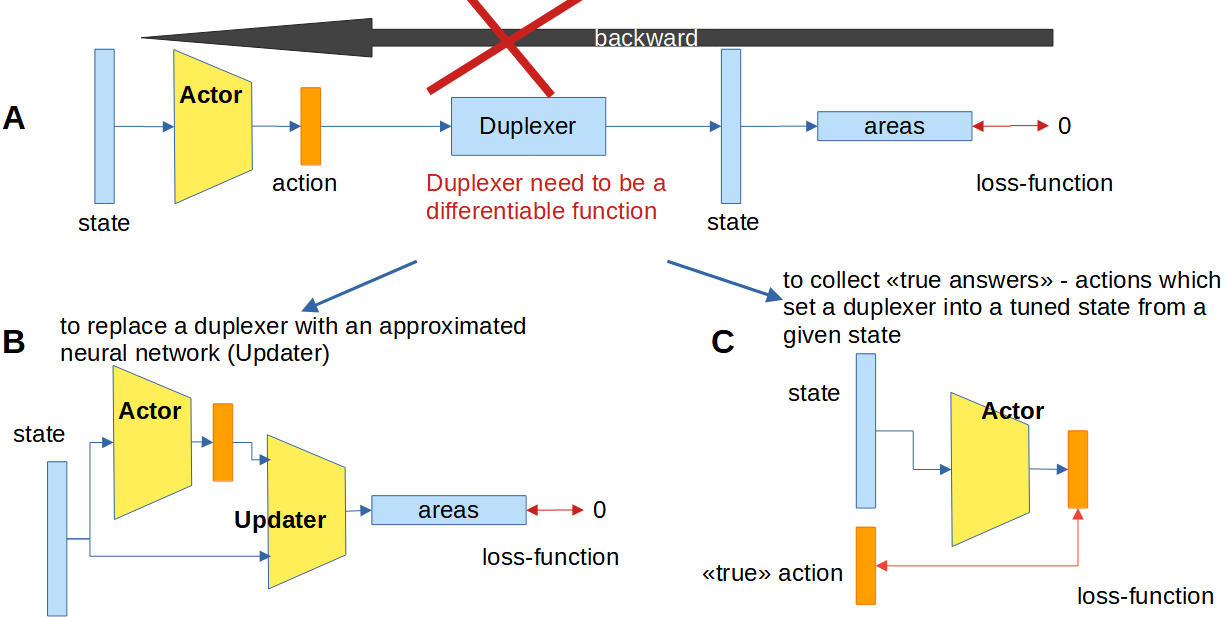} 
	\caption{The impossibility of training using a naive scheme (a) and ways to circumvent this limitation (b, c).}
	\label{fig:2}
\end{figure}

When training NN models for duplexer tuning, we would ideally use the areas (formula~\eqref{eq:1}) as the loss functions and backpropagate the gradients to the actor. However, the problem is that we cannot pass gradients from duplexer state to the actor, because a duplexer is a black box for us (see Figure~\ref{fig:2}a). The first way to bypass this limitation is to approximate a duplexer with some differentiable function, for example, another NN which we will refer to as ``Updater''. The second way is to collect a dataset with pairs duplexer state -- ``true action'' (an action that puts a duplexer into a tuned state, as we obtain data with simulator it is possible to collect such dataset). In the first variant our scheme contains two separate supervised learning tasks, in the second -- just one supervised learning task. We will discuss both ways in the next two subsections.

\subsection{Updater -- Actor scheme}
\label{upd}

The updater is neural network which mimics a duplexer (more precisely, the simulator) behavior. It takes an action vector and a current state (S11-, S21- and S31-curves) as an input and predicts new duplexer state after such action is applied. In principle, it is not necessary to predict the full state and it is sufficient to obtain areas discussed in~\ref{terms}.

To train the updater, a large number of triplets ``state'' -- ``action'' -- ``areas'' needs to be collected. Once the updater is trained, its weights should be frozen. The output of the actor is fed into the trained updater. The output of the updater network is compared to a target value using a loss function, and gradients are backpropagated through both networks to update the first one (actor), Figure~\ref{fig:2}b. Unfortunately, this strategy does not perform well. Even a small inaccuracy in the updater is likely to lead to wrong action suggested by the actor.

\subsection{Actor trained with ``true'' actions}
\label{true}

To perform this experiment, we fist collect a specific dataset in the following manner. First, screws of the duplexer are placed in random positions resulting in some state-tensor. Then, some initial action (obtained using a neural network or other algorithm) is performed, i.e. to rotate screws according this action-vector. As our environment keeps current screws positions and by design have an information about screws positions of the tuned duplexer (``golden positions''), we can calculate a ``true action'' -- an action which converts current state into the tuned state. It is just a difference between golden and actual positions. Hence, we can save pairs (current state -- true action) and train our model on such data in conventional supervised learning setup (Figure~\ref{fig:2}c).

We can choose a simple mean square error or mean absolute error as a loss function, but in this case, we will force actor to try to reach some defined positions of the screws. However, our task is not to place screws in specific positions since we just need to meet the requirements for S11. So, we manually have found a range of position for each screw around golden positions in which the requirements to the coefficients are still satisfied. Moreover, we have found that some screws have higher adjustment sensitivity. Therefore, the final loss function can be written as:

\begin{equation}
\text{loss} = \text{mean} (\text{ReLU}(\text{abs} (\text{predicted} - \text{golden}) - \delta) * k ), \label{eq:2}
\end{equation}

where $\delta$ is a permissible screw position deviation, $k$ are sensitivity coefficients (from 1.0 to 10.0 for different screws). This methodology gives much better results than the Updater-Actor scheme. Architectures of tested neural networks are described in the next subsection.

\subsection{Architectures}
\label{arch}

\begin{figure}[p]
    \centering
    \includegraphics[width=0.8\linewidth]{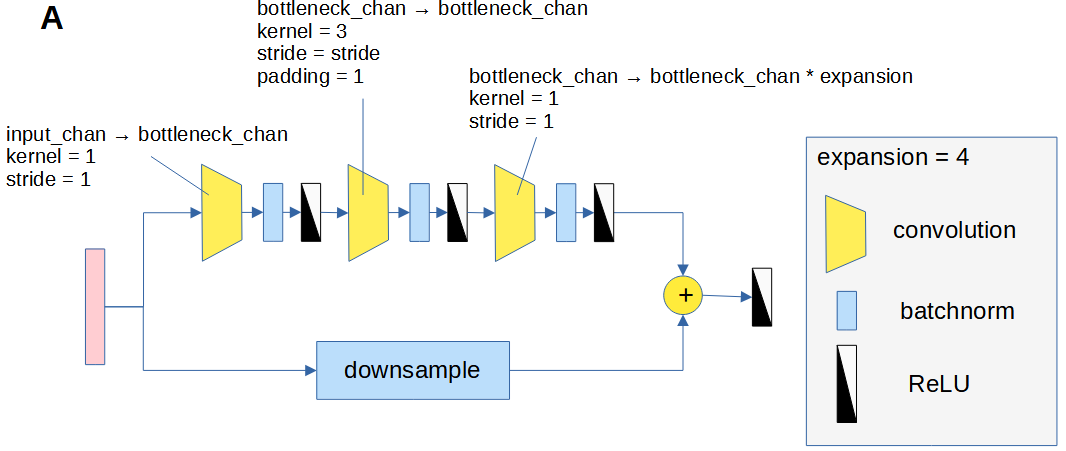}\\[1em]
    \includegraphics[width=0.8\linewidth]{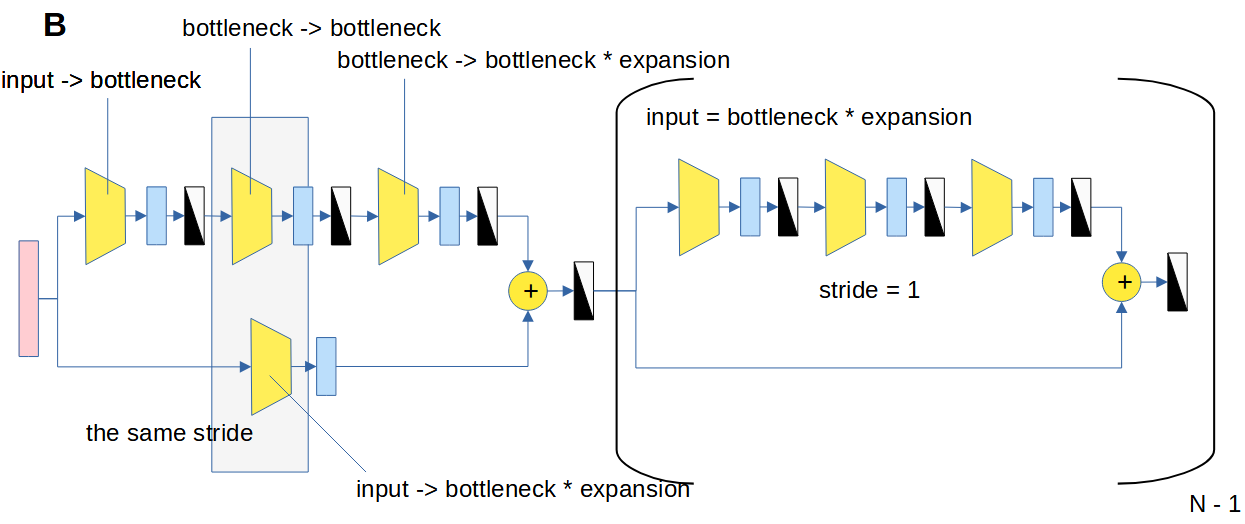}\\[1em]
    \includegraphics[width=0.8\linewidth]{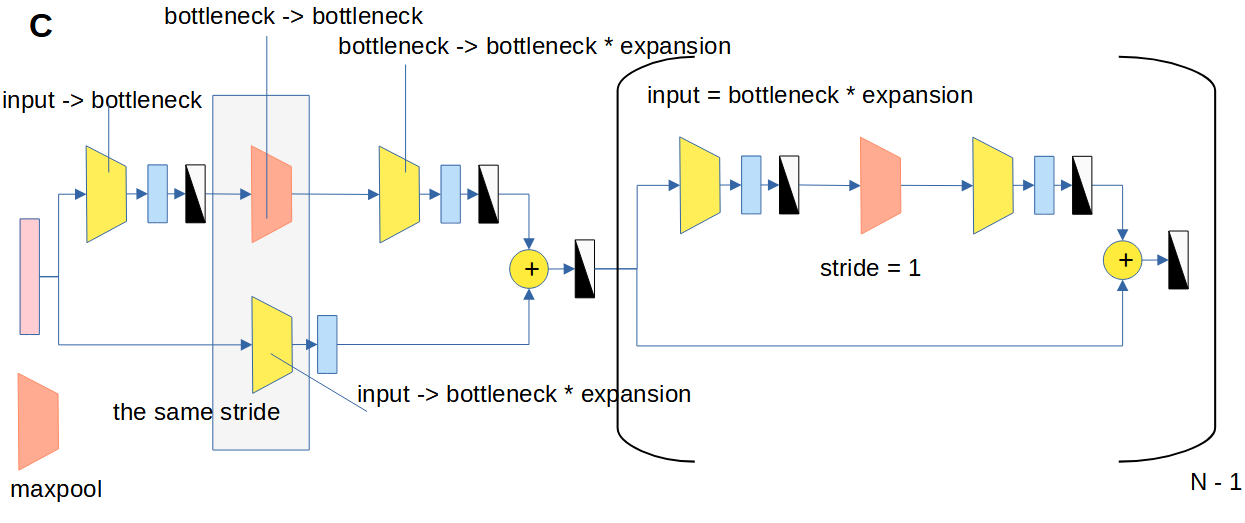}
    \caption{Elements of ResNet architecture: (a) a ResNet block which has 4 initialization parameters (the number of input channels, the number of bottleneck channels, the stride value and arbitrary neural network downsample (it may be empty)); (b) a ResNet layer with 4 parameters (the number of input channels, the number of bottleneck channels; the stride value and the number of blocks), note that the stride parameter applied only to the first block in the layer, other blocks have stride = 1, to fit tensor dimensionalities; (c) our variant of ResNet layer with some elements replaced with maxpooling operation.}
    \label{fig:3}
\end{figure}

Initially we start with very simple models as feed forward or convolution networks with some layers. Then, we continuously collected the dataset (pairs state -- ``true'' action) and adjusted the model's complexity to ensure it could adequately describe it. After the dataset exceeded 80000 records, simpler models showed significant errors. We then switched to a ResNet-like~\cite{12} architectures, which were able to handle even such a large dataset. Unlike the original ResNet which was designed for image processing and used 2d~convolutions, we use 1d~convolutions for S-curves. The main element of our architecture is 1d-ResNet-like block which consists of several convolutions, batch normalizations, ReLU activations and skip connections (see Figure~\ref{fig:3}a). These blocks are combined sequentially in ResNet-like layers (Figure~\ref{fig:3}b). We also used our own modification of ResNet layer, where one convolution is replaced with maxpooling (Figure~\ref{fig:3}c).

The total ResNet-like part used in all our experiments can be generally presented as a sequence of elements specified in Listing~\ref{lst:1}. The output of this sequence is passed through a feed forward network (which in the simplest case consists of one linear layer only) with output dimension equal to number of duplexer screws. Note that in case of fully tuned duplexer the actor output must return a zero-filled action vector. To enforce such behavior of the neural network, we introduce quantities that we refer as ``forces'' by analogy with driving forces. These forces are just areas (formula~\eqref{eq:1}) sliced into 20 subareas (10 per a passband) of equal frequency range and normalized to the range $[0;1]$. Each force has its own learnable coefficients for each screw. The forces are multiplied by their respective coefficients and summed to produce a single output value for each screw. It is equivalent to multiplication of forces-vector by a rectangular learnable matrix to produce output vector with dimensionality of the action. This result is element-wise multiplied by the output of the remaining part of the actor. Note, that forces are zero for the tuned duplexer and in this case the action-vector will be zeroed out.

\begin{lstlisting}[caption={The list of layers used in ResNet-like part of the actor. Conv is 1d-convolution layer, BN is batch norm layer, Layer is ResNet-like layer (Figure \protect\ref{fig:3}b), LayerMax is our modification of ResNet layer (Figure \protect\ref{fig:3}c). For ResNet-like layers the number of input and bottleneck channels are specified, N is a block number in ResNet-like layer.}, label={lst:1}]
NCHAN <- 64
Conv( 1 -> NCHAN, kernel=7, stride=2 )
BN
ReLU
Maxpool( kernel=3, stride=2 )
Layer( NCHAN -> NCHAN, N=3, stride=1 )
Layer( NCHAN * 4 -> NCHAN * 2, N=4, stride=2 )
Layer( NCHAN * 8 -> NCHAN * 4, N=6, stride=2 )
LayerMax( NCHAN * 16 -> NCHAN * 8, N=3, stride=2 )
AdaptiveMaxPool( 1 )
\end{lstlisting}

During further development of architectures, we concatenate some additional information to ResNet-like network output before feeding it into the fully connected network. For instance, it can be data about shape of S21 curve. The S21 curve looks like a plateau around 0~dB in high-frequency passband, with sharp drops on both sides of the plateau and gently sloping region in another passband. So, we can approximate these four regions of S21 using linear regression with the least-squares method and use the obtained linear regression coefficients (two for each of the four regions) as information about the shape of the curve. These 8~parameters are concatenated with the output of the ResNet-like part of the actor before being fed into the fully connected layers.

\subsection{Peak Encoders}
\label{peak}

\begin{figure}[h]
	\centering
	\includegraphics[width=\linewidth]{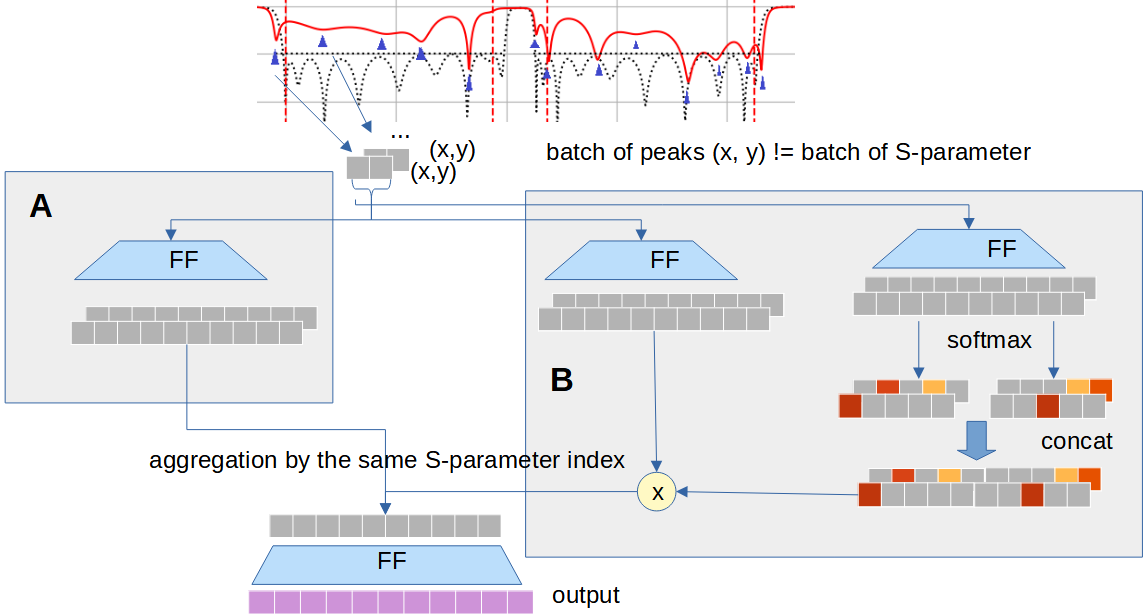} 
	\caption{Peak encoders: (a) simple -- just pass (x, y) of peaks into a feed forward neural network; (b) (x, y) pass into feed forward and (x) pass into another feed forward net and the result of subsequent softmax is used as mask for peak embedding. Both variants contain subsequent aggregation by S-parameter index. In the case (b) logits for softmax can be split on several parts (``heads'').}
	\label{fig:4}
\end{figure}

As typically S-curves look like a curve with some minima (negative peaks), we supposed that pre-calculated information about peaks' positions and amplitudes will improve the performance of the actor. To do this we collect indices of all peaks in all batches, and we pass pairs (peak index in tensor, amplitude of peak) through feed forward neural network to obtain a ``peak embedding''. As we have variable number of peaks per S-curve we aggregate (sum) embeddings of all peaks with the same batch index. Finally, aggregated embedding is passed through another feed forward network, as shown in Figure~\ref{fig:4}a. We will hereafter refer to such elements as peak encoders. The next variant of peak encoder includes softmax similar to attention mechanism in transformers~\cite{13}: the mask for softmax is obtained by passing peak position into feed forward network (we also tried to use pairs of positions and amplitudes, but the result was sufficiently worse), Figure~\ref{fig:4}b. Output of peak encoders was concatenated with ResNet-like network output.

\subsection{Solver}
\label{solv}

\begin{figure}[h]
	\centering
	\includegraphics[width=\linewidth]{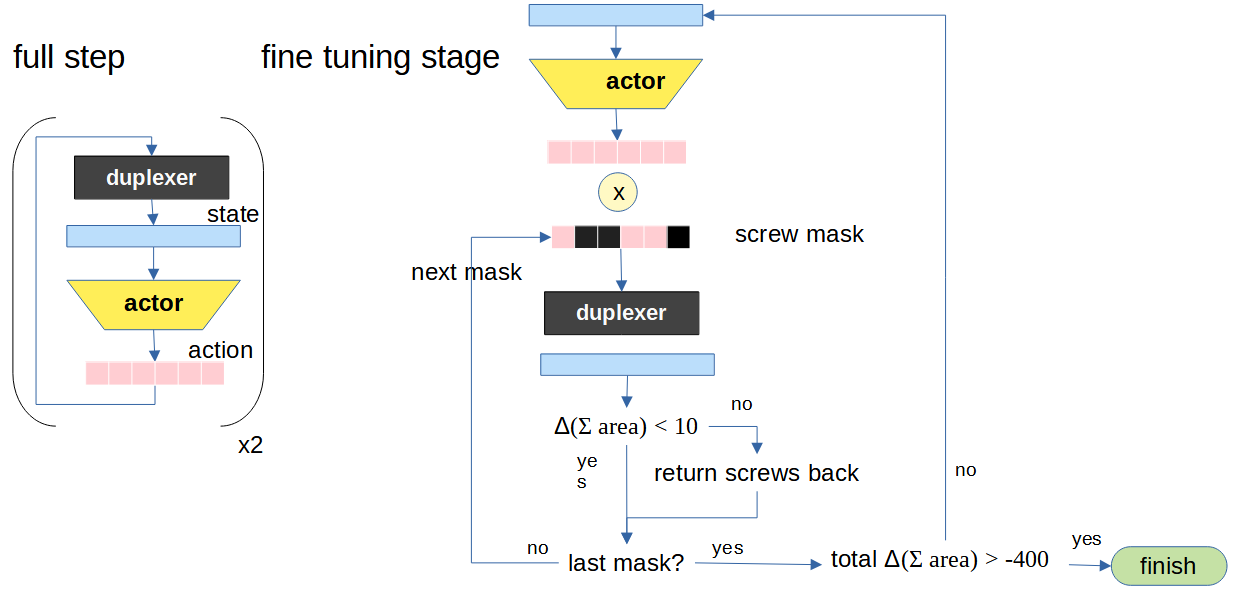} 
	\caption{The solver algorithm with several full steps and a fine-tuning stage.}
	\label{fig:5}
\end{figure}

According to our test, the value of loss function (formula~\eqref{eq:2}) below~0.01 is sufficient to reach the tuned state in a single action for a given dataset. However, in online settings we cannot obtain exactly the same states as we trained on, so the actor can only move to the state being closer to the desired state. But we can send a new state into the actor again and again. Unfortunately, after several such steps the environment state may start to degrade. For this reason, we introduce an external algorithm (referred as a ``solver'') that controls the actor. One of more robust algorithms of solver consist of 2 ``full'' steps (state $\rightarrow$ actor $\rightarrow$ action $\rightarrow$ environment $\rightarrow$ state) with a subsequent ``fine tuning stage''. Term ``full'' means that all screws are moving according to the actor output. The fine-tuning stage is carried out as follows: the actor calculates deltas for all screws, and they are randomly grouped into sets of~3 and applied batch by batch. If after applying batch of screws the metrics (area sum, formula~\eqref{eq:1}) degraded more than on 10~units the batch is canceled and these screws are returning back. The new action is calculated only after all screw batches have been tested. The fine-tuning step is repeated if the previous step has the total metric improvement greater then 400~units, see Figure~\ref{fig:5} for the algorithm. Due to mechanical wear of the screws, our purpose is to minimize the number of rotations of each screw, meaning that higher-quality actor riches the same (or better) result faster. For the same reason we are willing to accept even a screw batch which leads to a slight drop in the performance metric as otherwise we need to return them back and, consequently, do more rotations.

\section{Results}
\label{results}

\begin{figure}[t]
    \centering
    \begin{subfigure}{0.45\textwidth}
        \includegraphics[width=\linewidth]{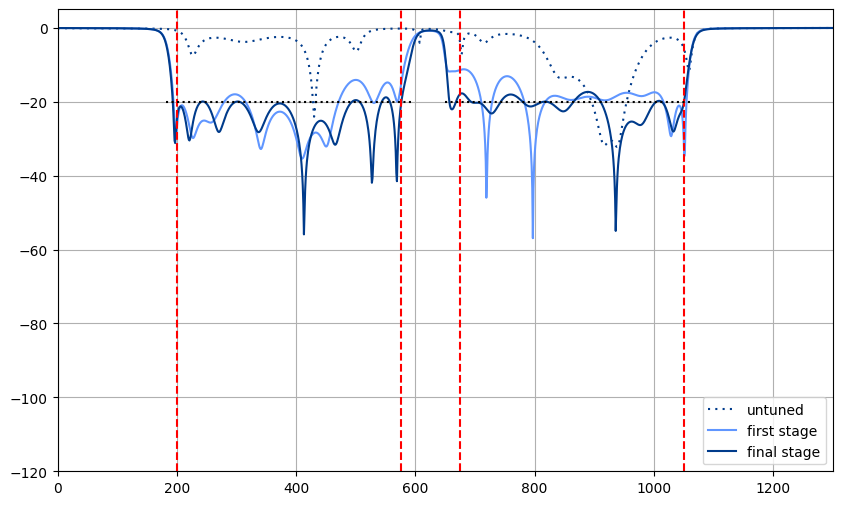}
        \label{fig:6a}
        \caption{}
    \end{subfigure}
    \hfill
    \begin{subfigure}{0.45\textwidth}
        \includegraphics[width=\linewidth]{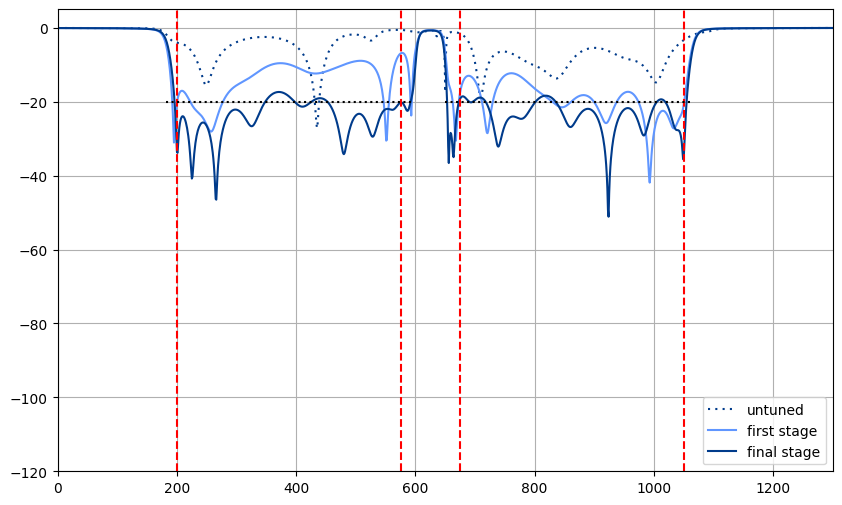}
        \label{fig:6b}
        \caption{}
    \end{subfigure}
    \caption{Result of solver work for different initial state obtained in online interaction with the simulator. Areas in high- and low frequency passbands: (a) 14.5 and 154.7; (b) 101.4 and 72.1.}
    \label{fig:6}
\end{figure}

Examples of evolution of S11-curve during tuning process from untuned state in online setting using solver are shown in Figure~\ref{fig:6}. The first stage shifts most part of S11 curve below -10~dB in both passbands and requires two rotations of each screw. The final stage requires a variable number of rotations, for cases shown in Figure~\ref{fig:6} this amounted totally to 167~(a) and 159~(b). As one can see the final state is very close to the requirements.

\begin{figure}[H]
	\centering
	\includegraphics[width=\linewidth]{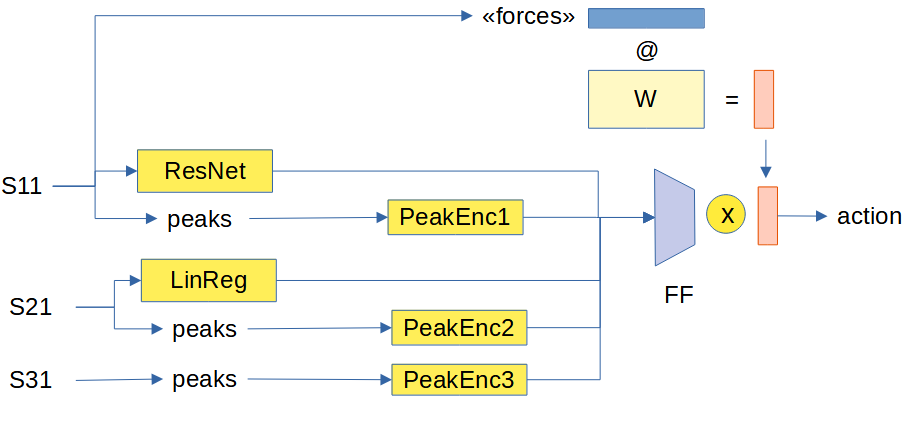} 
	\caption{General scheme for some architectures tested here.}
	\label{fig:7}
\end{figure}

\subsection{Generalization}
\label{gen}

Even the largest dataset collected as part of this study is much smaller than all possible states of a duplexer. So, it is very important that the actor gives a low error even on data which was not used during training. The ability to have good values of loss function on another dataset (not used in training) is referred to as generalization ability or generalization. For our purposes, additional data collected from the simulator can be used as another dataset. We collect one dataset near untuned state and two others as partially tuned. The loss value on them can be considered as a measure of generalization. The results of generalization study for different actors (Figure~\ref{fig:7}) are summarized in Table~\ref{tab:1}. All actors listed in Table~\ref{tab:1} have very low loss values on the training dataset, much lower than~0.01. First significant improvement of generalization is observed when peaks encoders are added not only for S11, but also for S21 and S31 (actor22 \textrightarrow 23). As we know high number of parameters can lead to overfitting, so a dependence of generalization on parameter number must have a maximum. Here we see such behavior for channels number in convolution filters (actors 24, 25, 25ch36). Starting from actor 25s2 we increased the stride size in the first block of the first ResNet layer. This leads not only to improvements in performance but also to the reduction in size of tensors and, consequently, required size. Adding coefficients of linear regression (slope and free term) for 4~regions of S21 parameter to final feed forward network also notably improves the quality of the actor (25s2 \textrightarrow 26). For unknown reasons, adding similar information from S31 made the metrics worse. Finally, a little bit of improvement was reached by replaceing simple peak encoder for S11 to the one with softmax (see Figure~\ref{fig:4}).

\begin{table}[H]
    \centering
    \caption{The comparison of generalization for some architectures tested here. NCHAN is the number of channels after the first convolution (see Listing \protect\ref{lst:1}); BLOCKS is a list of block numbers in each ResNet layer (see Listing \protect\ref{lst:1}).}
    \label{tab:1}
    \begin{tabular}{|c|c|c|c|c|}
        \hline
        Actor \# & Description & gen 1 & gen 2 & gen 3 \\
        \hline\hline
        22 & \makecell[l]{NCHAN = 64; BLOCKS = \lbrack 3,4,6,3 \rbrack\\No Linear Regression\\PeakEncoder1 = Simple\\PeakEncoder2, 3 = None} & 0.8 & \textasciitilde0.9 & \textasciitilde0.9 \\
        \hline
        23 & \makecell[l]{NCHAN = 64; BLOCKS = \lbrack 3,4,6,3 \rbrack\\No Linear Regression\\PeakEncoder1,2,3 = Simple} & 0.76 & \textasciitilde0.19 & \textasciitilde0.19 \\
        \hline
        24 & \makecell[l]{NCHAN = 64; BLOCKS = \lbrack 3,4,5,2 \rbrack\\No Linear Regression\\PeakEncoder1,2,3 = Simple} & 0.76 & \textasciitilde0.14 & \textasciitilde0.14 \\
        \hline
        25 & \makecell[l]{NCHAN = 50; BLOCKS = \lbrack 3,4,5,2 \rbrack\\No Linear Regression\\PeakEncoder1,2,3 = Simple} & 0.65 & 0.127 & 0.106 \\
        \hline
        25ch36 & \makecell[l]{NCHAN = 36; BLOCKS = \lbrack 3,4,5,2 \rbrack\\No Linear Regression\\PeakEncoder1,2,3 = Simple} & 0.70 & 0.138 & 0.117 \\
        \hline
        25s2 & \textsuperscript{*}layer1: stride 1\textrightarrow2 & 0.66 & 0.126 & 0.105 \\
        \hline
        26 & \makecell[l]{NCHAN = 50; BLOCKS = \lbrack 3,4,5,2 \rbrack\\PeakEncoder1,2,3 = Simple} & 0.630 & 0.122 & 0.099 \\
        \hline
        26soft & \makecell[l]{NCHAN = 50; BLOCKS = \lbrack 3,4,5,2 \rbrack\\PeakEncoder1 = Softmax\\PeakEncoder2,3 = Simple} & 0.628 & 0.122 & 0.097 \\
        \hline
    \end{tabular}
    \vspace{1mm}
    \parbox{\linewidth}{\raggedright\textsuperscript{*} This and subsequent models have stride=2 in the first ResNet-like.}
\end{table}

\section{Conclusion}
\label{concl}

In this work we developed a neural network model for the cavity duplexer tuning. The model (actor) uses S-curves as an input and returns a vector, each element of which indicates how much the corresponding adjustment screw should be rotated. Since the actor is unable to bring the duplexer to a tuned state in a single iteration, a sequence of several steps guided by an additional algorithm can calibrate the device very close to the desired state. The most robust architectures of the actors include processing of S-curves with deep 1d-convolution networks and incorporating high-structured information like peaks positions and amplitudes and slopes of S-curves in specific regions. Finally, we can reach almost the tuned state within 4-5~rotations per screw.

\section*{Acknowledgement}
\label{ackn}

The author would like to thank his colleagues Dr Nikita Gabdullin and Vasily Dolmatov for their advice and fruitful discussion.

\label{references}
\bibliographystyle{IEEEtran}
\bibliography{IEEEabrv,ms}

\end{document}